\documentclass{article}

    \PassOptionsToPackage{numbers, compress}{natbib}

    \usepackage[final]{neurips_2024}

\usepackage[utf8]{inputenc} 
\usepackage[T1]{fontenc}    
\usepackage{hyperref}       
\usepackage{url}            
\usepackage{booktabs}       
\usepackage{amsfonts}       
\usepackage{nicefrac}       
\usepackage{microtype}      
\usepackage{xcolor}         
\usepackage{graphicx}
\usepackage{amsmath}
\usepackage{threeparttable}
\usepackage{multirow}

\title{MonoMAE: Enhancing Monocular 3D Detection through Depth-Aware Masked Autoencoders}

\author{%
  Xueying Jiang$^{1}$, Sheng Jin$^{1}$, Xiaoqin Zhang$^{2}$, Ling Shao$^{3}$, Shijian Lu$^{1}$\thanks{Corresponding author.} \\
  $^{1}$S-Lab, Nanyang Technological University, Singapore\\
  $^{2}$College of Computer Science and Technology, Zhejiang University of Technology, China\\
  $^{3}$UCAS-Terminus AI Lab, University of Chinese Academy of Sciences, China
}

\begin{document}

\maketitle

\begin{abstract}
Monocular 3D object detection aims for precise 3D localization and identification of objects from a single-view image. Despite its recent progress, it often struggles while handling pervasive object occlusions that tend to complicate and degrade the prediction of object dimensions, depths, and orientations. We design MonoMAE, a monocular 3D detector inspired by Masked Autoencoders that addresses the object occlusion issue by masking and reconstructing objects in the feature space. MonoMAE consists of two novel designs. The first is depth-aware masking that selectively masks certain parts of non-occluded object queries in the feature space for simulating occluded object queries for network training. It masks non-occluded object queries by balancing the masked and preserved query portions adaptively according to the depth information. The second is lightweight query completion that works with the depth-aware masking to learn to reconstruct and complete the masked object queries. With the proposed feature-space occlusion and completion, MonoMAE learns enriched 3D representations that achieve superior monocular 3D detection performance qualitatively and quantitatively for both occluded and non-occluded objects. Additionally, MonoMAE learns generalizable representations that can work well in new domains.
\end{abstract}

\section{Introduction}
\label{sec:intro}
3D object detection has emerged as one key component in various navigation tasks such as autonomous driving, robot patrolling, etc. Compared with prior studies relying on LiDAR~\cite{zhou2018voxelnet, lang2019pointpillars, yin2021center} or multi-view images~\cite{li2022bevformer, liu2022petr, yang2023bevformer}, monocular 3D object detection offers a more cost-effective and accessible alternative which identifies objects and predicts their 3D locations from single-view images. On the other hand, monocular 3D object detection is much more challenging due to the lack of 3D information from multi-view images or LiDAR data.

\begin{figure*}[htbp]
    \centering
    \includegraphics[width=\linewidth]{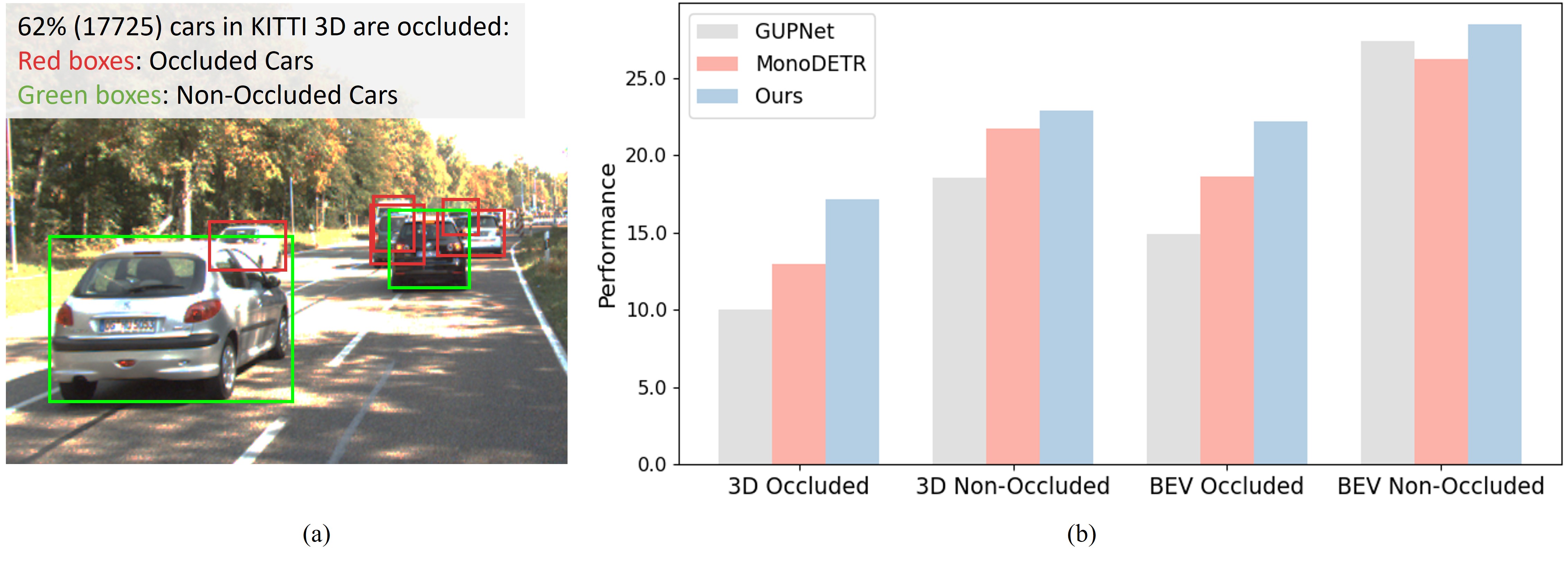}

    \caption{
    Object occlusion is pervasive and affects monocular 3D detection: Object occlusion is pervasive, e.g., 62\% (17725) cars in the KITTI 3D dataset suffer from various occlusions as illustrated in (a). Prevalent monocular 3D detection techniques such as GUPNet~\cite{lu2021geometry} and MonoDETR~\cite{zhang2023monodetr} are clearly affected by object occlusions in both 3D space (3D) and the bird’s eye view (BEV) space as in (b). 
    The proposed MonoMAE simulates and learns object occlusions by feature masking and completing which improves detection consistently for both occluded and non-occluded objects.
    }
    \label{fig:motivation}
\end{figure*}

Among various new challenges in monocular 3D detection, object occlusion, which exists widely in natural images as illustrated in Figure~\ref{fig:motivation} (a), becomes a critical issue while predicting 3D locations in terms of object depths, object dimensions, and object orientations. Most existing monocular 3D detectors such as MonoDETR\cite{zhang2023monodetr} and GUPNet~\cite{lu2021geometry} neglect the object occlusion issue which demonstrates clear performance degradation as illustrated in Figure~\ref{fig:motivation} (b). A simple idea is to learn to reconstruct the occluded object regions whereby occluded objects can be handled similarly as non-occluded objects. On the other hand, reconstructing occluded object regions in the image space is complicated due to the super-rich variation of object occlusions in scene images.

Inspired by the Masked Autoencoders (MAE)~\cite{he2022masked} that randomly occludes image patches and reconstructs them in representation learning, we treat object occlusions as natural masking and train networks to complete occluded object regions to learn occlusion-tolerant representations. To this end, we design MonoMAE, a novel monocular 3D detection framework that adopts the idea of MAE by first masking certain object regions in the feature space (for simulating object occlusions) and then reconstructing the masked object features (for learning occlusion-tolerant representations). MonoMAE consists of a depth-aware masking module and a lightweight completion network. The depth-aware masking simulates object occlusions by masking the features of non-occluded objects adaptively according to the object depth information. It generates pairs of non-occluded and masked (i.e., occluded) object representations that can be directly applied to train the lightweight completion network, aiming for completing the occluded objects and learning occlusion-tolerant representations. Note that MonoMAE introduces little computational overhead in inference time as it requires no object masking in the inference stage.

The contributions of this work can be summarized in three major aspects. \textit{First}, we design MonoMAE, a MAE-inspired monocular 3D detection framework that tackles object occlusions effectively by masking and reconstructing object regions in the feature space. To the best of our knowledge, this is the first work that explores masking-reconstructing for the task of monocular 3D object detection.
\textit{Second}, we design adaptive image masking and a lightweight completion network that mask non-occluded objects adaptively according to the object depth (for simulating object occlusions) and reconstruct the masked object regions (for learning occlusion-tolerant representations), respectively. \textit{Third}, extensive experiments over KITTI 3D and nuScenes show that MonoMAE outperforms the state-of-the-art consistently and it can generalize to new domains as well.

\section{Related Work}
\label{sec:related_work}

\subsection{Monocular 3D Object Detection}

Monocular 3D detection aims for the identification and 3D localization of objects from a single-view image. Most existing work can be broadly classified into two categories. The first employs convolutional neural networks, where most methods follow conventional 2D detectors~\cite{duan2019centernet, jin2024llms}. The standard approach learns monocular 3D detectors from single-view images only~\cite{he2019mono3d,brazil2019m3d, chen2020monopair, zhou2021monoef, zhang2023monodetr, liu2020smoke}. In addition, several studies explore to leverage extra training data, such as LiDAR point clouds~\cite{ma2019accurate,wang2019pseudo, chen2021monorun,reading2021categorical,peng2024learning, peng2022did}, depth maps~\cite{ding2020learning,qin2021monogrnet,peng2022did,jiang2024weakly, ma2020rethinking}, and 3D CAD models~\cite{chen2016monocular,liu2021autoshape,murthy2017reconstructing} to acquire more depth information. Beyond that, several studies exploit the geometry relation between 2D and 3D spaces in different ways. For example, M3D-RPN~\cite{brazil2019m3d} applies the powerful 2D detector FPN~\cite{ren2015faster} for 3D detection. MonoDLE~\cite{ma2021delving} aligns the centers of 2D and 3D boxes for better 3D localization. GUPNet~\cite{lu2021geometry} leverages uncertainty modeling to estimate the height of 3D boxes from the 2D boxes.

The second introduces powerful visual transformers~\cite{zhu2021deformable, jiang2023modify, carion2020end, zhang2023black} for more accurate monocular 3D detection ~\cite{huang2022monodtr, zhang2023monodetr, zhou2023monoatt, wu2023attention, wu2023monopgc}. For example, MonoDTR~\cite{huang2022monodtr} integrates context- and depth-aware features and injects depth positional hints into transformers. MonoDETR~\cite{zhang2023monodetr} modifies the transformer to be depth-aware and guides the detection process by contextual depth cues. However, most existing methods neglect object occlusions that exist widely in natural images and often degrade the performance of monocular 3D object detection clearly. We adopt the transformer architecture to learn occlusion-tolerant representations that can handle object occlusion effectively without requiring any extra training data or annotations.

\subsection{Occlusions in 3D Object Detection}

Object occlusion is pervasive in scene images and it has been investigated in several 2D and 3D vision tasks~\cite{yao2023occlusion, su2023opa, chu2022visibility, li2022a3d, li2016amodal, li2023gin, li2023muva}. One typical approach learns to estimate the complete localization of occluded objects. For example, Mono-3DT~\cite{hu2019joint} estimates complete 3D bounding boxes by re-identifying occluded vehicles from a sequence of 2D images. BtcDet~\cite{xu2022behind} leverages object shape priors to learns to estimate the complete shapes of partially occluded objects. Several studies consider the degree of occlusions in training. For example, MonoPair~\cite{chen2020monopair} exploits the relation of paired samples and encodes spatial constraints of occluded objects from their neighbors. HMF~\cite{liu2022fine} introduces an anti-occlusion loss to focus on occluded samples. Different from existing methods, the proposed MonoMAE learns enriched and occlusion-tolerant representations by masking and completing object parts in the feature space.

\subsection{Masked Autoencoders in 3D Tasks}
Masked Autoencoders (MAE)~\cite{he2022masked} learn visual representations by masking image patches and reconstructing them, and it has been explored in several point cloud pre-training studies. For outdoor point cloud pre-training, Occupancy-MAE~\cite{min2023occupancy} exploits range-aware random masking that employs three masking levels to deal with the sparse voxel occupancy structures of LiDAR point clouds. GD-MAE~\cite{yang2023gd} introduces a Generative Decoder to merge the surrounding context to restore the masked tokens hierarchically. For indoor point cloud pre-training, Point-MAE~\cite{pang2022masked} adopts MAE to directly reconstruct the 3D coordinates of masked tokens. I2P-MAE~\cite{zhang2023learning} introduces 2D pre-trained models, and it enhances 3D pre-training with diverse 2D semantics. PiMAE~\cite{chen2023pimae} learns cross-modal representations with MAE by interactively handling point clouds and RGB images. Different from existing studies, the proposed MonoMAE handles monocular 3D detection from single-view images and it focuses on object occlusions by learning to complete occluded object regions in the feature space.

\section{Proposed Method}
\label{sec:method}

\subsection{Problem Definition}
Monocular 3D detection takes a single RGB image as input, aiming to classify objects and predict their 3D bounding boxes. The prediction of each object is composed of the object category $C$, a 2D bounding box $B_{2D}$, and a 3D bounding box $B_{3D}$. The 3D bounding box $B_{3D}$ can be decomposed to the object 3D location $(x_{3D}, y_{3D}, z_{3D})$, the object dimensions in object height, width and length $(h_{3D}, w_{3D}, l_{3D})$, as well as the object orientation $\theta$.

\begin{figure*}[t]
    \centering
    \includegraphics[width=0.95\linewidth]{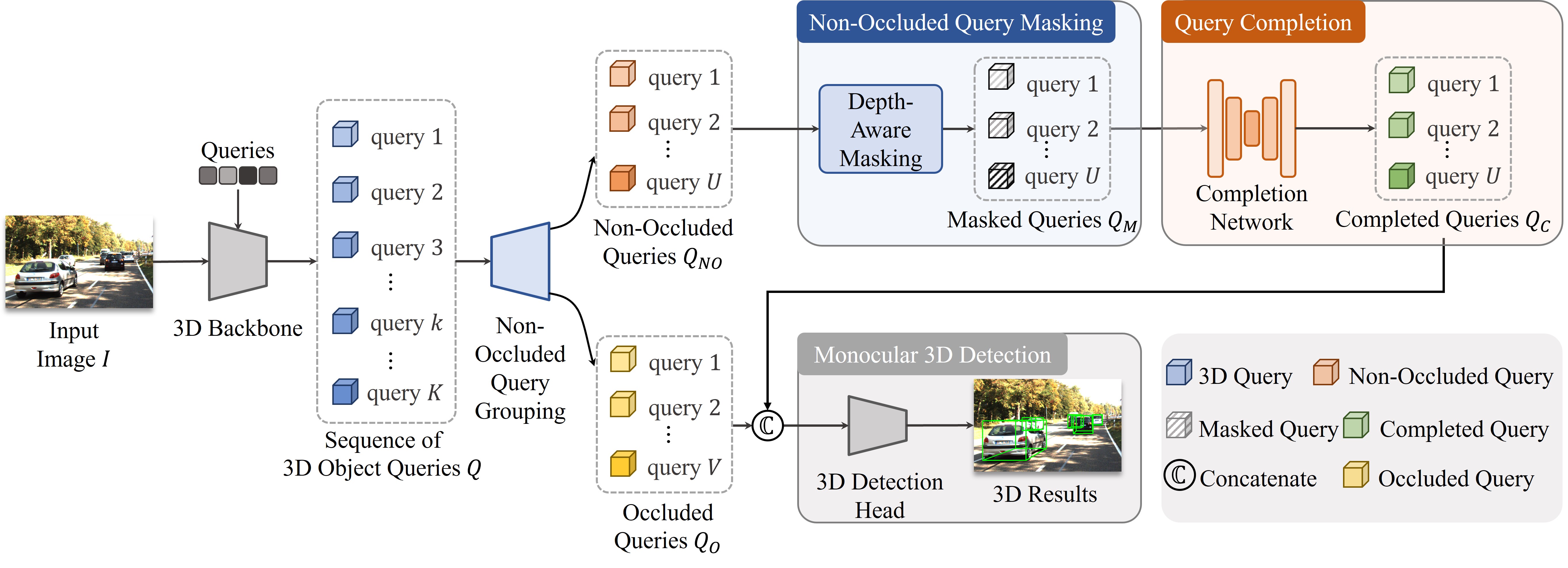}

    \caption{
    The framework of MonoMAE training: Given a single-view image, the 3D Backbone extracts 3D object query features which are grouped into non-occluded query features and occluded query features by the Non-Occluded Query Grouping. The Depth-Aware Masking then masks the non-occluded query features to simulate object occlusions adaptively based on the object depth, and the Completion Network then learns to reconstruct the masked queries. Finally, the completed and the occluded query features are concatenated to train the 3D Detection Head for 3D predictions.
    }
    \label{fig:overall_architecture}
\end{figure*}

\subsection{Overall Framework}
Figure~\ref{fig:overall_architecture} shows the framework of the proposed MonoMAE. Given an input image $I$, the 3D Backbone first generates a sequence of 3D object queries $Q =[q_{1}^{}, q_{2}^{}, \cdots, q_{K}^{}]$ ($K$ denotes query number), and the Non-Occluded Query Grouping then classifies the queries into two groups including non-occluded queries $Q^{NO} =[q_{1}^{NO}, q_{2}^{NO}, \cdots, q_{U}^{NO}]$ and occluded queries $Q^{O} =[q_{1}^{O}, q_{2}^{O}, \cdots, q_{V}^{O}]$ ($U$ and $V$ are the number of non-occluded and occluded queries). The Non-Occluded Query Masking then masks $Q^{NO}$ to produce masked queries according to their depth $D =[d_{1}^{}, d_{2}^{}, \cdots, d_{U}^{}]$, leading to the masked queries $Q^{M} =[q_{1}^{M}, q_{2}^{M}, \cdots, q_{U}^{M}]$. The Query Completion further reconstructs $Q^{M}$ to produce the completed queries $Q^{C} =[q_{1}^{C}, q_{2}^{C}, \cdots, q_{U}^{C}]$. Finally, the occluded queries $Q^{O}$ and the completed queries $Q^{C}$ are concatenated and fed to the Monocular 3D Detection for 3D detection predictions. Note the inference does not involve the Non-Occluded Query Masking, and it just concatenates the completion of occluded queries $Q^{O}$ (i.e., $Q^{C}$) with the non-occluded queries $Q^{NO}$ and feeds the concatenated queries to the 3D Detection Head for 3D predictions.

\subsection{Non-Occluded Query Masking}
\label{sec:non_occluded_query_masking}

Queries predicted by the 3D Backbone are either occluded or non-occluded, depending on whether the corresponding objects are occluded in the input image. In MonoMAE, we mask the non-occluded queries in the feature space to simulate occlusions, aiming to generate pairs of non-occluded and masked (i.e., occluded) queries for learning occlusion-tolerant object representations.

Specifically, we design a Non-Occluded Query Grouping module to identify non-occluded queries and then feed them into a Depth-Aware Masking module to synthesize occlusions, with more detail to be elaborated in the following subsections.

\textbf{Non-Occluded Query Grouping.}
The Non-Occluded Query Grouping classifies the queries based on whether their corresponding objects are occluded or non-occluded. With no information about whether the input queries are occluded, we design an occlusion classification network $\mathrm{\Phi}_{\text{O}}$ to predict the occlusion conditions $O^p=[o_{1}^{p}, o_{2}^{p}, \cdots, o_{K}^{p}]$ of queries $Q =[q_{1}^{}, q_{2}^{}, \cdots, q_{K}^{}]$, where for the $i$-th query $o_{i}^{p} = \mathrm{\Phi}_{\text{O}} (q_{i}^{})$. The Non-Occluded Query Grouping can be formulated by:

\begin{equation}
    \left\{\begin{array}{ll}
q_{i} \in Q^{NO} & \text { if } o_{i}^{p} = 0\\  
q_{i} \in Q^{O} & \text { if } o_{i}^{p} = 1
\end{array}\right. ,
\end{equation}
where $o_{i}^{p} = 0$ denotes the query is non-occluded, and $o_{i}^{p} = 1$ denotes the query is occluded. 
The occlusion classification network is trained with the occlusion classification loss $L_{occ}$ as follows:
\begin{equation}
    L_{occ}=CE(O^p, O^{gt}),  
    \label{equation_L_occ}
\end{equation}
where $CE$ is the $Cross$ $Entropy$ loss. We adopted the bipartite matching~\cite{carion2020end} to match the predicted queries and objects in the image, where only matched queries have ground truth $O^{gt}$ of KITTI 3D~\cite{geiger2012we} about whether they are occluded or not. 

\begin{figure*}[t]
    \centering
    \includegraphics[width=0.92\linewidth]{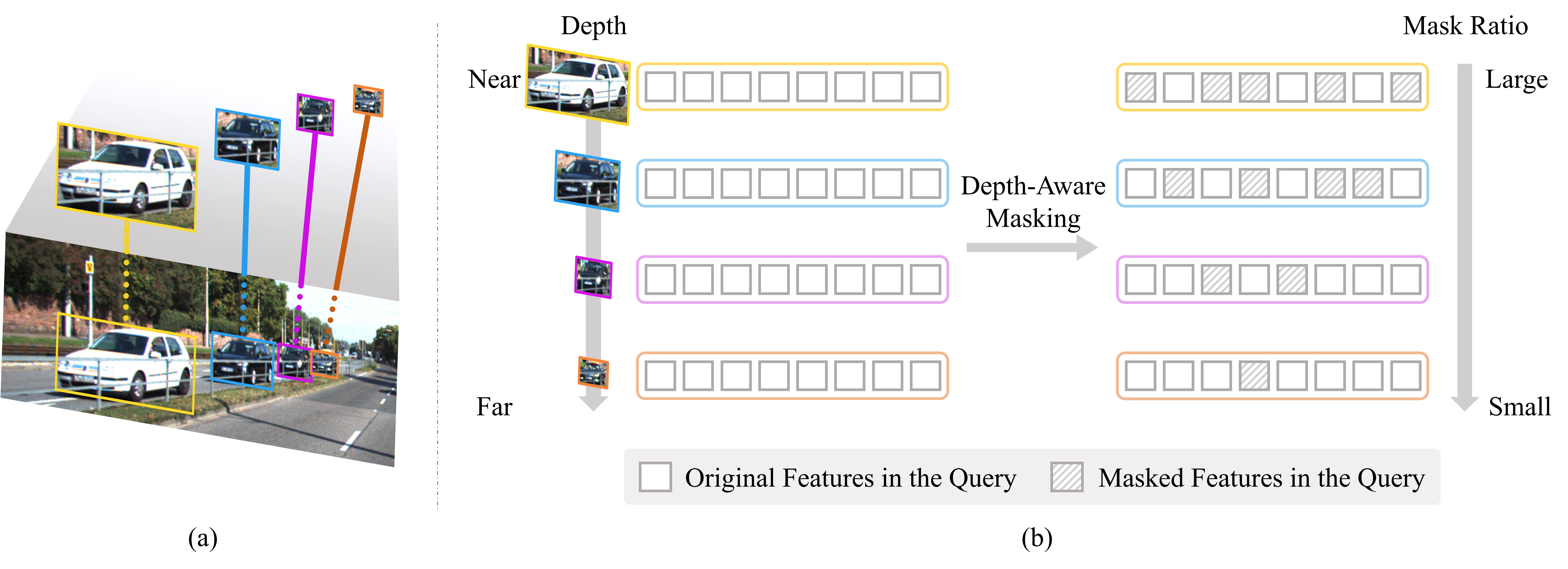}
    \caption{
    Illustration of the Depth-Aware Masking. (a) Objects farther away are usually smaller capturing less visual information. (b) The Depth-Aware Masking determines the mask ratio of an object according to its depth - the closer the object is, the larger the mask ratio is applied, thereby compensating the information deficiency for objects that have larger distances from the camera.
    }
    \label{fig:depth_aware_masking}
\end{figure*}

\textbf{Depth-Aware Masking.} 
We design depth-aware masking to adaptively mask non-occluded query features to simulate occlusions in the feature space, aiming to create non-occluded and occluded (i.e., masked) pairs for learning occlusion-tolerant representations. As illustrated in Figure~\ref{fig:depth_aware_masking}, the depth-aware masking determines the mask ratio according to the object depth - the closer the object, the larger the mask ratio, thereby compensating the information deficiency of distant objects. In addition, we simulate occlusions by masking in the feature space, as masking and reconstructing at the image level is complicated and computationally intensive.

The depth-aware masking first obtains the query depth before query masking. Without backward gradient propagation, it adopts the 3D Detection Head to obtain the depth $D^{}=[d_{1}^{}, d_{2}^{}, \cdots, d_{U}^{}]$ for non-occluded queries. With the predicted depth, each non-occluded query is randomly masked as illustrated in Figure~\ref{fig:depth_aware_masking}. Specifically, objects that are more distant from the camera are usually captured with less visual information. The depth-aware masking accommodates this by assigning a smaller masking ratio to them, thereby keeping more visual information for distant objects for proper visual representation learning.

The mask ratio $r$ of each query is determined by:
\begin{equation}
    r = 1.0 - d_{i}/D_{max},
    \label{equation_R_mask}
\end{equation}
where $r$ is the applied mask ratio for each query,
$d_{i}$ is the depth for the $i$-th query, and $D_{max}$ is the maximum depth in datasets. The masks $M^{}=[m_{1}^{}, m_{2}^{}, \cdots, m_{U}^{}]$ generated for queries obey a Bernoulli Distribution.

Finally, the query masking is formulated by:
\begin{equation}
    q_{i}^{M} = q_{i}^{NO} * m_{i},
\end{equation}
where $q_{i}^{M}$ is the masked query, $q_{i}^{NO}$ is the non-occluded query, and $m_{i}$ is the generated mask.

\subsection{Query Completion}
\label{sec:query_completion}
The query completion learns to reconstruct the adaptively masked queries, aiming to produce completed queries whereby the network learns occlusion-tolerant representations that are helpful in detecting occluded objects. We design a completion network $\mathrm{\Phi}_{\text{C}}$ to reconstruct the masked queries. The Completion Network has an hourglass structure consisting of three conv-bn-relu blocks and one conv-bn block for 3D query completion. 
The completed query $q_{i}^{C}$ is obtained by:

\begin{equation}
    q_{i}^{C} = \mathrm{\Phi}_{\text{C}} (q_{i}^{M}),
\end{equation}
where $q_{i}^{M}$ is the masked query. 
The Completion Network is trained under the supervision of the non-occluded queries before masking, where a completion loss $L_{com}$ is formulated as follows:

\begin{equation}
    L_{com} = L_1^s(Q^{NO}, Q^{C}),
    \label{equation_L_com}
\end{equation}
where $L_1^s$ denotes the SmoothL1 loss~\cite{girshick2015fast}, $Q^{NO}$ denotes the non-occluded queries, and $Q^{C}$ denotes the queries completed by the Completion Network.

\begin{table*}[t]
\caption{
    Benchmarking on the KITTI 3D \textit{test} set. All experiments adopt AP$|_{R_{40}}$ metric with an IoU threshold of 0.7. Best in \textbf{bold}, second \underline{underlined}.
}
\centering
\renewcommand\arraystretch{1.0}
\setlength\tabcolsep{1pt}
\scalebox{0.99}{
\begin{threeparttable}
\begin{tabular}{l|c|c|ccc|ccc}
\hline
\multirow{2}{*}{Method} & \multirow{2}{*}{Venue} & \multirow{2}{*}{Extra Data}  & \multicolumn{3}{c|}{AP$_{3D}   ($IoU$=0.7)|_{R_{40}}$} &  \multicolumn{3}{c}{AP$_{BEV}   ($IoU$=0.7)|_{R_{40}}$} \\ 
           &             &                                                      & Easy                 & Moderate            & Hard    & Easy                 & Moderate            & Hard             \\ \hline\hline
MonoRUn~\cite{chen2021monorun}   &    CVPR 21        & \multirow{5}{*}{LiDAR}
 &   19.65      & 12.30         &   10.58    & 27.94   & 17.34   & 15.24  \\
MonoDTR~\cite{huang2022monodtr} & CVPR 22 &          & 21.99
 &   15.39      &  12.73        &  28.59     &  20.38  &   17.14   \\
MonoDistill~\cite{chong2022monodistill} &  ICLR 22            & 
 &  22.97       &   16.03       & 13.60      &  31.87  &   22.59 &  19.72 \\
DID-M3D~\cite{peng2022did}   &    ECCV 22        & &   24.40      &   16.29       &   13.75    & 32.95   &   22.76 & 19.83  \\
MonoNeRD~\cite{xu2023mononerd}  &  ICCV 23           & 
 &  22.75       &   \underline{17.13}       &  \underline{15.63}     & 31.13   &  \underline{23.46} & \underline{20.97}  \\
\hline
D4LCN~\cite{ding2020learning}   & CVPR 20    & \multirow{3}{*}{Depth}      & 16.65
 &   11.72      &  9.51        &    22.51   &   16.02 &  12.55     \\
DDMP-3D~\cite{wang2021depth}   &  CVPR 21    &         & 19.71
 & 12.78        &  9.80        &  28.08     &  17.89  &  13.44    \\
DD3D~\cite{park2021pseudo} &  ICCV 21    &  & 23.22  & 16.34        &  14.20       &  30.98    &  22.56  &  20.03     \\
\hline
Kinematic3D~\cite{brazil2020kinematic}      &   ECCV 20      & Video
 &  19.07       &   12.72       &  9.17     &  26.69  & 17.52   &  13.10 \\
\hline
AutoShape~\cite{liu2021autoshape}  &   ICCV 21          & CAD
 &   22.47      &   14.17       &   11.36    &  30.66  &  20.08  &  15.59 \\
\hline
MonoFlex~\cite{zhang2021objects}  &    CVPR 21         & \multirow{9}{*}{None}
 &  19.94       &    13.89      &   12.07    & 28.23   &  19.75  & 16.89  \\
MonoRCNN~\cite{shi2021geometry} &   ICCV 21           & 
 &  18.36       &      12.65    &  10.03     &  25.48  & 18.11   &  14.10 \\
GUPNet~\cite{lu2021geometry}  &  ICCV 21    &       & 20.11
 &  14.20       &  11.77        &   -    &  -    &   -    \\
DEVIANT~\cite{kumar2022deviant}    &   ECCV 22        & 
 &  21.88       &     14.46     &  11.89     & 29.65   & 20.44   &   17.43 \\
MonoCon~\cite{liu2022learning} &   AAAI 22           & 
 &  22.50       &  16.46        & 13.95      & 31.12   & 22.10   & 19.00  \\
MonoDETR~\cite{zhang2023monodetr}  &    ICCV 23         & 
 &   25.00     &   16.47       &  13.58     &  \underline{33.60}  & 22.11   & 18.60  \\
MonoUNI~\cite{jia2023monouni} & NeurIPS 23  &  & 24.75  & 
  16.73  & 13.49   &   -    &  -    &   -  \\
MonoCD~\cite{yan2024monocd} & CVPR 24  &   & \underline{25.53} & 16.59  & 14.53
 & 33.41 &   22.81  & 19.57           \\
\hline
\textbf{Ours}     &    -      & None
 &      \textbf{25.60} & \textbf{18.84} & \textbf{16.78} & \textbf{34.15} &  \textbf{24.93} &  \textbf{21.76}  \\
\hline

\end{tabular}

\end{threeparttable}
}

\label{tab:table_1_kitti_test_car_0.7}
\end{table*}

\subsection{Loss Functions}

The overall objective consists of three losses including $L_{occ}$, $L_{com}$, and $L_{base}$ where $L_{occ}$ and $L_{com}$ are defined in Equation~\ref{equation_L_occ} and Equation~\ref{equation_L_com}, and $L_{base}$ denote losses for supervising the 3D box predictions.
Specifically, $L_{base}$ includes losses for supervising the 3D box predictions including each object's 3D locations, height, width, length and orientation. We set the weight for each loss item to 1.0, and the overall loss function is formulated as follows:
\begin{equation}
    L = L_{occ} + L_{com} + L_{base}.
\end{equation}

\section{Experiments}
\label{sec:exp}

\subsection{Experimental Settings}

\textbf{Datasets.}
We benchmark our method over two public datasets in monocular 3D object detection.

$\bullet$ KITTI 3D~\cite{geiger2012we} comprises 7,481 training images and 7,518 testing images, with training-data labels publicly available and test-data labels stored on a test server for evaluation. Following~\cite{chen2016monocular}, we divide the 7,481 training samples into a new train set with 3,712 images and a validation set with 3,769 images for ablation studies. 

$\bullet$ NuScenes~\cite{caesar2020nuscenes} comprises 1,000 video scenes, including RGB images captured by 6 surround-view cameras. The dataset is split into a training set (700 scenes), a validation set (150 scenes), and a test set (150 scenes). Following~\cite{brazil2019m3d, shi2021geometry, lu2021geometry, kumar2022deviant, jia2023monouni}, the performance on the validation set of nuScenes is reported.

In addition, we perform evaluations on the most representative Car category of KITTI 3D and nuScenes datasets as in prior studies~\cite{simonelli2019disentangling, shi2021geometry, wang2021fcos3d, zhang2023monodetr}

\textbf{Evaluation Metrics.}
For KITTI 3D, we follow~\cite{simonelli2019disentangling} and adopt AP$|_{R_{40}}$, the average of the AP of 40 recall points as the evaluation metric. We report the average precision on BEV and 3D object detection by AP$_{BEV}|_{R_{40}}$ and AP$_{3D}|_{R_{40}}$ with a threshold of 0.7 for both test and validation sets. For the nuScenes dataset, we adopt the mean absolute depth errors~\cite{shi2021geometry} in evaluations.


\textbf{Implementation Details.}
We conduct experiments on one NVIDIA V100 GPU and train the framework for 200 epochs with a batch size of 16 and a learning rate of $2 \times 10^{-4}$. We use the AdamW~\cite{loshchilov2018decoupled} optimizer with weight decay $10^{-4}$. We employ ResNet-50~\cite{he2016deep} as the Transformer-based backbone and adopt the 3D detection head from~\cite{zhang2023monodetr} as our detection framework.


\subsection{Benchmarking with the State-of-the-Art} 
We benchmark MonoMAE with state-of-the-art monocular 3D object detection methods both quantitatively and qualitatively.

\textbf{Quantitative Benchmarking.}
Table~\ref{tab:table_1_kitti_test_car_0.7} shows quantitative experiments on the test set of dataset KITTI 3D, where all evaluations were performed on the official online test server~\cite{geiger2012we} for fairness. We can see that MonoMAE achieves superior detection performance consistently across all metrics, without using any extra training data such as image depths, video sequences, LiDAR points, and CAD 3D models. In addition, MonoMAE outperforms more for the Moderate and Hard categories where various occlusions happen much more frequently than the Easy category. The superior performance is largely attributed to our designed depth-aware masking and completion network, which masks queries to simulate object occlusions in the feature space and reconstructs the masked queries to learn occlusion-tolerant visual representations, respectively.

\begin{figure*}[t!]
    \centering
    \includegraphics[width=0.98\linewidth]{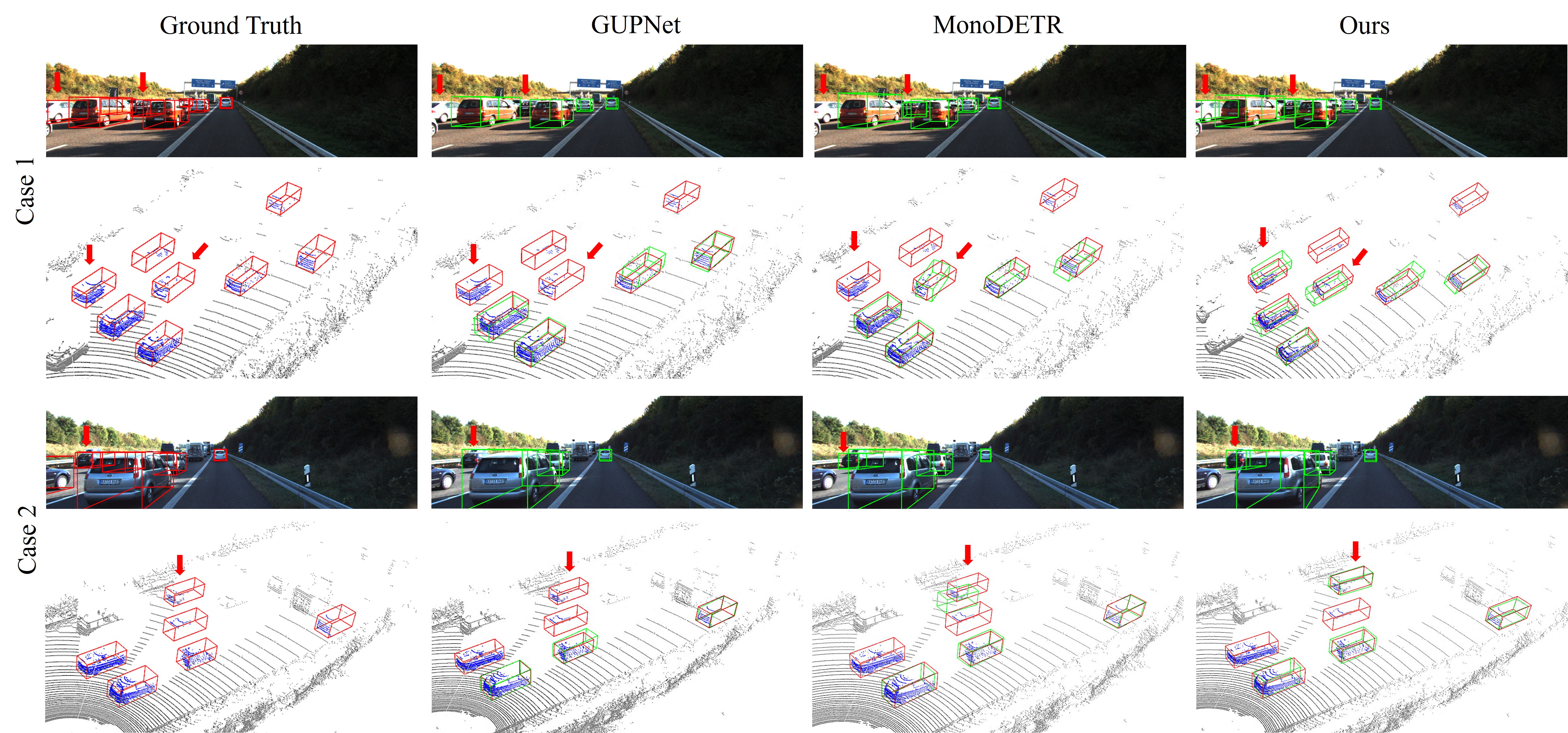}

    \caption{
    Detection visualization over the KITTI \textit{val} set. Ground-truth annotations are highlighted by \textcolor{red}{red} boxes, and predictions by MonoMAE and two state-of-the-art methods are highlighted by \textcolor{green}{green} boxes. Red arrows highlight objects that have very different predictions across the compared methods. The ground truth of LiDAR point clouds is provided for visualization only, and they are not used in MonoMAE training. Best viewed in color and zoom-in.
    }
    \label{fig:visualization}
\end{figure*}


\begin{table*}[t]
\caption{
    Ablation study of technical designs in MonoMAE on the KITTI 3D \textit{val} set. `NOQG’, `DAM’, and `CN’ denote Non-Occluded Query Grouping, Depth-Aware Masking, and Completion Network, respectively. The symbol * indicates the baseline. The best results are highlighted in \textbf{bold}.
}
\renewcommand\arraystretch{1.0}
\setlength\tabcolsep{2pt}
\centering
\scalebox{1.0}{
\begin{threeparttable}
\begin{tabular}{c|ccc|ccc|ccc}
\hline
\multirow{2}{*}{Index} & \multirow{2}{*}{NOQG} & \multirow{2}{*}{DAM} & \multirow{2}{*}{CN} & \multicolumn{3}{c|}{AP$_{3D}   ($IoU$=0.7)|_{R_{40}}$} &  \multicolumn{3}{c}{AP$_{BEV}   ($IoU$=0.7)|_{R_{40}}$} \\ 
                        &    &    &                                                   & Easy                 & Moderate            & Hard    & Easy                 & Moderate            & Hard             \\ \hline\hline
1*    &  \checkmark    &    &     &  24.85  &  16.21 &  14.74   &  34.53 &  23.99  & 18.84 \\
2    &      &  \checkmark  &     &  23.33  & 15.09  &  13.20      & 32.68 & 21.80 & 17.43  \\
3    &    &    & \checkmark    &   27.33 &  18.52 & 14.95   &  36.51 &  24.21 &  19.12  \\
4    &  \checkmark    &  \checkmark  &   & 24.69 &  15.71  &  13.57        &  33.46 &  23.03  &  18.19 \\
5    &  \checkmark    &    &  \checkmark   & 27.25 & 18.76 &  15.45 & 36.81 &  25.18 & 20.05   \\
6    &    & \checkmark   &  \checkmark   &  28.39       &   19.35       &  15.87     & 37.59 & 26.27 & 21.33 \\
\hline
7    &  \checkmark    & \checkmark     &   \checkmark    &     \textbf{30.29}    &   \textbf{20.90}       & \textbf{17.61}      & \textbf{40.26}   & \textbf{27.08}   & \textbf{23.14}  \\
\hline
\end{tabular}
\end{threeparttable}
}

\label{tab:table_2_different_module}
\end{table*}

\begin{table*}[t]
\caption{
    Ablation study of masking strategies on the KITTI 3D \textit{val} set. The best results are in \textbf{bold}.
}
\renewcommand\arraystretch{1.0}
\setlength\tabcolsep{1pt}
\centering
\scalebox{0.99}{
\begin{threeparttable}
\begin{tabular}{c|c|ccc|ccc}
\hline
\multirow{2}{*}{Index} & \multirow{2}{*}{Masking Strategy}  & \multicolumn{3}{c|}{AP$_{3D}   ($IoU$=0.7)|_{R_{40}}$} &  \multicolumn{3}{c}{AP$_{BEV}   ($IoU$=0.7)|_{R_{40}}$} \\ 
                        &                  & Easy                 & Moderate            & Hard    & Easy                 & Moderate            & Hard             \\ \hline\hline
1    &    Image Masking      &  20.51       & 15.03         &   13.24    &  27.76  & 19.74   & 16.71  \\
2    &    Query Masking (w/o Depth-Aware) &     27.14      &  18.47        &  15.02   &  36.98  & 25.52   & 20.64  \\
\hline
3    &   Query Masking (w/ Depth-Aware)       &   \textbf{30.29}    &   \textbf{20.90}       & \textbf{17.61}      & \textbf{40.26}   & \textbf{27.08}   & \textbf{23.14} \\
\hline
\end{tabular}
\end{threeparttable}
}

\label{tab:table_3_different_mask_approach}
\end{table*}

\textbf{Qualitative Benchmarking.} 
Figure~\ref{fig:visualization} shows qualitative benchmarking on the KITTI 3D val set. It can be observed that compared with two state-of-the-art methods GUPNet and MonoDETR, the proposed MonoMAE produces more accurate 3D detection consistently for both non-occluded and occluded objects, even for challenging scenarios like distant objects. 
Specifically, GUPNet and MonoDETR tend to miss the detection of highly occluded object in Cases 1 and 2 as highlighted by red arrows. As a comparison, MonoMAE performs clearly better by detecting those challenging objects successfully, demonstrating its superior capability on handling object occlusions.

\subsection{Ablation Study}
We conduct extensive ablation studies to examine the proposed MonoMAE. Specifically, we examine MonoMAE from the aspect of the technical designs, query masking strategies, as well as loss functions.


\textbf{Network Designs.} 
We examine the effectiveness of two key designs in MonoMAE, namely, the Depth-Aware Masking module (DAM) and the Completion Network (CN) (on the validation set of KITTI 3D), as shown in Table~\ref{tab:table_2_different_module}. We formulate the baseline by including the Non-Occluded Query Grouping module (NOQG), which does not affect the network training as both identified occluded and non-occluded queries are fed to train 3D detectors. When CN is not used in Rows 2 and 4, the 3D detection degrades as queries are masked but not reconstructed which leads to further information loss. While not incorporating DAM in Rows 3 and 5, the detection improves clearly compared with the baseline, as the completion helps learn better representations for naturally occluded queries. In addition, incorporating DAM and CN on top of NOQG in Row 7 performs clearly better than incorporating DAM and CN alone in Row 6, as the former applies masking and completion to non-occluded queries only. It also shows that masking naturally occluded queries to train the completion network is harmful to the learned representations.

\textbf{Masking Strategies.} 
We examine how different masking strategies affect monocular 3D detection. We studied three masking strategies as shown in Table~\ref{tab:table_3_different_mask_approach}. The first strategy masks the \textit{input images} randomly, aiming to assess the value of masking and completing in the feature instead of image space. We can observe that the image-level masking yields clearly lower performance as compared with query masking in the feature space, largely due to the complication in masking and reconstructing images with a lightweight completion network. The second strategy masks query features randomly without considering object depths, aiming to evaluate the importance of object depths in query masking. 
The experiments show that random query masking outperforms the image-level masking significantly.
The third strategy performs the proposed depth-aware query masking. It outperforms the feature-space random masking consistently, demonstrating the value of object depths for query masking.

\begin{table*}[t]
\caption{
    Ablation study of the loss functions on the KITTI 3D \textit{val} set. $L_{occ}$ and $L_{com}$ refer to the occlusion classification loss and the completion loss, respectively. The best results are in \textbf{bold}.
}
\centering
\renewcommand\arraystretch{1.0}
\setlength\tabcolsep{4pt}
\scalebox{1.0}{
\begin{threeparttable}
\begin{tabular}{c|cc|ccc|ccc}
\hline
\multirow{2}{*}{Index} & \multirow{2}{*}{$L_{occ}$} & \multirow{2}{*}{$L_{com}$} & \multicolumn{3}{c|}{AP$_{3D}   ($IoU$=0.7)|_{R_{40}}$} &  \multicolumn{3}{c}{AP$_{BEV}   ($IoU$=0.7)|_{R_{40}}$} \\ 
                        &                           &                           & Easy                 & Moderate            & Hard    & Easy                 & Moderate            & Hard             \\ \hline\hline
1              &   \checkmark  &     &  28.37  &  19.61  & 16.01       &  37.48 &  26.55 &  21.50   \\
2              &     &   \checkmark  &       26.36  &  19.15  & 15.88   &  36.76   &  26.49   &  22.62 \\
\hline
3              &  \checkmark   & \checkmark    &   \textbf{30.29}    &   \textbf{20.90}       & \textbf{17.61}      & \textbf{40.26}   & \textbf{27.08}   & \textbf{23.14}  \\
 
\hline
\end{tabular}
\end{threeparttable}
}

\label{tab:table_4_loss}
\end{table*}

\textbf{Loss Functions.} 
We examine the impact of the occlusion classification loss $L_{occ}$ and the completion loss $L_{com}$ in Equations~\ref{equation_L_occ} and~\ref{equation_L_com}, where $L_{occ}$ supervises the occlusion classification network (in Non-Occluded Query Grouping) to predict whether the queries are occluded and $L_{com}$ supervises the Completion Network to reconstruct the masked queries. As Table~\ref{tab:table_4_loss} shows, while implementing $L_{occ}$ alone, the occlusion prediction is supervised while the query reconstruction is unsupervised. The network under such an objective does not learn well as the Completion Network cannot reconstruct object queries well without sufficient supervision. While implementing $L_{com}$ alone, the occlusion classification network cannot identify occluded and non-occluded queries accurately where many occluded queries are fed for masking, leading to more query occlusion and poor detection performance. While employing both losses concurrently, the performance improves significantly as non-occluded queries can be identified for masking and reconstruction, leading to occlusion-tolerant representations.

\subsection{Discussions}

\begin{table*}[t]
\caption{
    Comparison on inference speed of several monocular 3D detection methods. 
    Ours* denotes the proposed MonoMAE without including the Completion Network.
}
\centering
\renewcommand\arraystretch{1.0}
\setlength\tabcolsep{2.9pt}
\scalebox{0.98}{
\begin{tabular}{l|c|c|c|c|c}
\hline
Method    & GUPNet~\cite{lu2021geometry}  & MonoDTR~\cite{huang2022monodtr}
&    MonoDETR~\cite{zhang2023monodetr}                   & Ours*  & Ours\\ \hline\hline
Inference Time (ms) & 40 & 37 & 43 & 36  & 38 \\  \hline
\end{tabular}
}

\label{tab:table_6_inference_time}
\end{table*}

\begin{table*}[t]
\caption{
    Cross-dataset evaluations that perform training on the KITTI train set, and testing on the KITTI val and nuScenes val sets. We adopt the evaluation metric mean absolute error of the depth ($\downarrow$). Best is highlighted in \textbf{bold}, and second \underline{underlined}.
}
\centering
\renewcommand\arraystretch{1.0}
\setlength\tabcolsep{3pt}
\scalebox{0.99}{
\begin{threeparttable}
\begin{tabular}{l|cccc|cccc}
\hline
\multirow{2}{*}{Method}  & \multicolumn{4}{c|}{KITTI Val} &  \multicolumn{4}{c}{nuScenes frontal Val} \\ 
& 0-20                 & 20-40            & 40-$\infty$  & All  &  0-20                 & 20-40            & 40-$\infty$  & All       \\ \hline\hline
M3D-RPN~\cite{brazil2019m3d}   &   0.56  &  1.33 &  2.73 &  1.26         & 0.94 & 3.06  & 10.36 &  2.67   \\
MonoRCNN~\cite{shi2021geometry}   &  0.46  & 1.27 &  2.59  & 1.14 &  0.94 &  2.84 &  8.65  & 2.39    \\
GUPNet~\cite{lu2021geometry}   &  0.45  & 1.10  & 1.85 &  0.89  & 0.82 & 1.70 &  6.20 &  1.45   \\
DEVIANT~\cite{kumar2022deviant}   &  0.40 &  1.09  & 1.80 &  \underline{0.87} &  0.76 &  \underline{1.60}
&  \textbf{4.50} &  \textbf{1.26}    \\
MonoUNI~\cite{jia2023monouni}   &   \underline{0.38}  & \underline{0.92}  & \underline{1.79} &  \underline{0.87} &  \underline{0.72} &  1.79 &  4.98 &  1.43  \\
\hline
\textbf{MonoMAE (Ours)}  &    \textbf{0.36}     &  \textbf{0.91}   &  \textbf{1.74}   &   \textbf{0.86}    &  \textbf{0.71}  & \textbf{1.57} & \underline{4.95}  & \underline{1.40}   \\

\hline
\end{tabular}
\end{threeparttable}
}

\label{tab:table_7_nuscenes}
\end{table*}

\textbf{Efficiency Comparison.} 
We compare the inference time of several representative monocular 3D detection methods on the KITTI val set, where all compared methods are evaluated with one NVIDIA V100 GPU under the same computational environment for fairness. As Table~\ref{tab:table_6_inference_time} shows, GUPNet, MonoDTR, and MonoDETR have an average inference time of 40ms, 37ms, and 43ms for each image, respectively. As a comparison, the proposed MonoMAE takes the shortest inference time, demonstrating its good efficiency in monocular 3D detection. 
Further, we analyzed the Completion Network in terms of network parameters and floating-point operations per second (FLOPs), showing it has very limited 2.22G parameters and 0.08M in FLOPs.

\textbf{Generalization Ability.}
We examine the generalization capability of the proposed MonoMAE by directly applying the KITTI-trained MonoMAE model to the car Category of the nuScenes validation set without additional training. The detection performance on the KITTI validation set is also reported for reference.
Table~\ref{tab:table_7_nuscenes} shows that MonoMAE attains the highest or second-highest detection performance across various metrics on the nuScenes frontal validation set. This indicates that despite the domain shift from KITTI to nuScenes, MonoMAE still maintains satisfactory performance. 
Since DEVIANT~\cite{kumar2022deviant} is equivariant to the depth translations, it sometimes has higher performance.


\section{Conclusion}
\label{sec:conclusion}

This paper presents MonoMAE, a novel method inspired by the Masked Autoencoders (MAE) to deal with the pervasive occlusion problem in the monocular 3D object detection task. MonoMAE consists of two key designs. The first is a depth-aware masking module, which simulates the occlusion for non-occluded object queries in the feature space during training. The second is a lightweight completion network, which reconstructs and completes the masked object queries.
Quantitative and qualitative experiment results show that MonoMAE learns enhanced 3D representations and achieves superior monocular 3D detection performance for both occluded and non-occluded objects. Moving forward, we plan to investigate generative approaches to simulate natural occlusion patterns for various 3D detection tasks. 

\textbf{Limitations.} MonoMAE leverages depth-aware masking to mask non-occluded queries to simulate object occlusions in the feature space. However, the masked queries may have different patterns as compared with the features of naturally occluded object queries. 
Such a gap could affect the reconstruction of complete queries and monocular 3D detection performance. 
This issue could be mitigated by introducing generative networks that learn distributions from extensive real-world data for generating occlusion patterns that are more similar to natural occlusions.

\begin{ack}
This study is supported under the RIE2020 Industry Alignment Fund – Industry Collaboration Projects (IAF-ICP) Funding Initiative, as well as cash and in-kind contribution from the industry partner(s).
\end{ack}


\bibliographystyle{plain}
\small\bibliography{egbib}

\clearpage

\appendix

\end{document}